\documentclass[11pt,a4paper]{article}
\usepackage[hyperref]{acl2020}
\usepackage{times}
\usepackage{latexsym}
\usepackage{booktabs}       
\usepackage{amsfonts}       
\usepackage{nicefrac}
\usepackage{latexsym}
\usepackage{multirow}
\usepackage{booktabs}
\usepackage{amsmath}
\usepackage{booktabs}

\usepackage{graphicx}
\usepackage{subfigure}

\newtheorem{Def}{Definition}

\newcommand*\samethanks[1][\value{footnote}]{\footnotemark[#1]}

\usepackage{microtype}

\aclfinalcopy 
\def\aclpaperid{3068} 

\title{Extractive Summarization as Text Matching}

\author{Ming Zhong\thanks{\ \  These two authors contributed equally.}, Pengfei Liu\samethanks, Yiran Chen, Danqing Wang, Xipeng Qiu\thanks{\ \  Corresponding author.} , Xuanjing Huang \\
  Shanghai Key Laboratory of Intelligent Information Processing, Fudan University \\
  School of Computer Science, Fudan University \\
  825 Zhangheng Road, Shanghai, China \\
  \texttt{\{mzhong18,pfliu14,yrchen19,dqwang18,xpqiu,xjhuang\}@fudan.edu.cn}
  }

\date{}

\begin{document}

\maketitle

\begin{abstract}



This paper creates a \emph{paradigm shift} with regard to the way we build neural extractive summarization systems. Instead of following the commonly used framework of extracting sentences individually and modeling the relationship between sentences, we formulate the extractive summarization task as a \emph{semantic text matching} problem, in which a source document and candidate summaries will be (extracted from the original text) matched in a semantic space. Notably, this paradigm shift to semantic matching framework is \emph{well-grounded} in our comprehensive analysis of the inherent gap between sentence-level and summary-level extractors based on the property of the dataset.

Besides, even instantiating the framework with a simple form of a matching model, we have driven the state-of-the-art extractive result on CNN/DailyMail to a new level (44.41 in ROUGE-1). Experiments on the other five datasets also show the effectiveness of the matching framework.  We believe the power of this matching-based summarization framework has not been fully exploited. To encourage more instantiations in the future,  we have released our codes, processed dataset, as well as generated summaries in {\url{https://github.com/maszhongming/MatchSum}}.

\end{abstract}

\section{Introduction}

The task of automatic text summarization aims to compress a textual document to a shorter highlight while keeping salient information on the original text.
In this paper, we focus on extractive summarization since it usually generates semantically and grammatically correct sentences \cite{dong2018banditsum,nallapati2017summarunner} and computes faster.



Currently, most of the neural extractive summarization systems score and extract sentences (or smaller semantic unit \cite{xu2019discourse}) one by one from the original text, model the relationship between the sentences, and then select several sentences to form a summary. \citet{cheng2016neural, nallapati2017summarunner} formulate the extractive summarization task as a sequence labeling problem and solve it with an encoder-decoder framework. These models make independent binary decisions for each sentence, resulting in high redundancy. A natural way to address the above problem is to introduce an auto-regressive decoder \cite{chen2018fast, jadhav2018extractive, zhou2018neural}, allowing the scoring operations of different sentences to influence on each other. Trigram Blocking \cite{paulus2017deep,liu2019text}, as a more popular method recently, has the same motivation. At the stage of selecting sentences to form a summary, it will skip the sentence that has trigram overlapping with the previously selected sentences. Surprisingly, this simple method of removing duplication brings a remarkable performance improvement on CNN/DailyMail.

\begin{figure}
    \centering
    \includegraphics[width=0.8\linewidth]{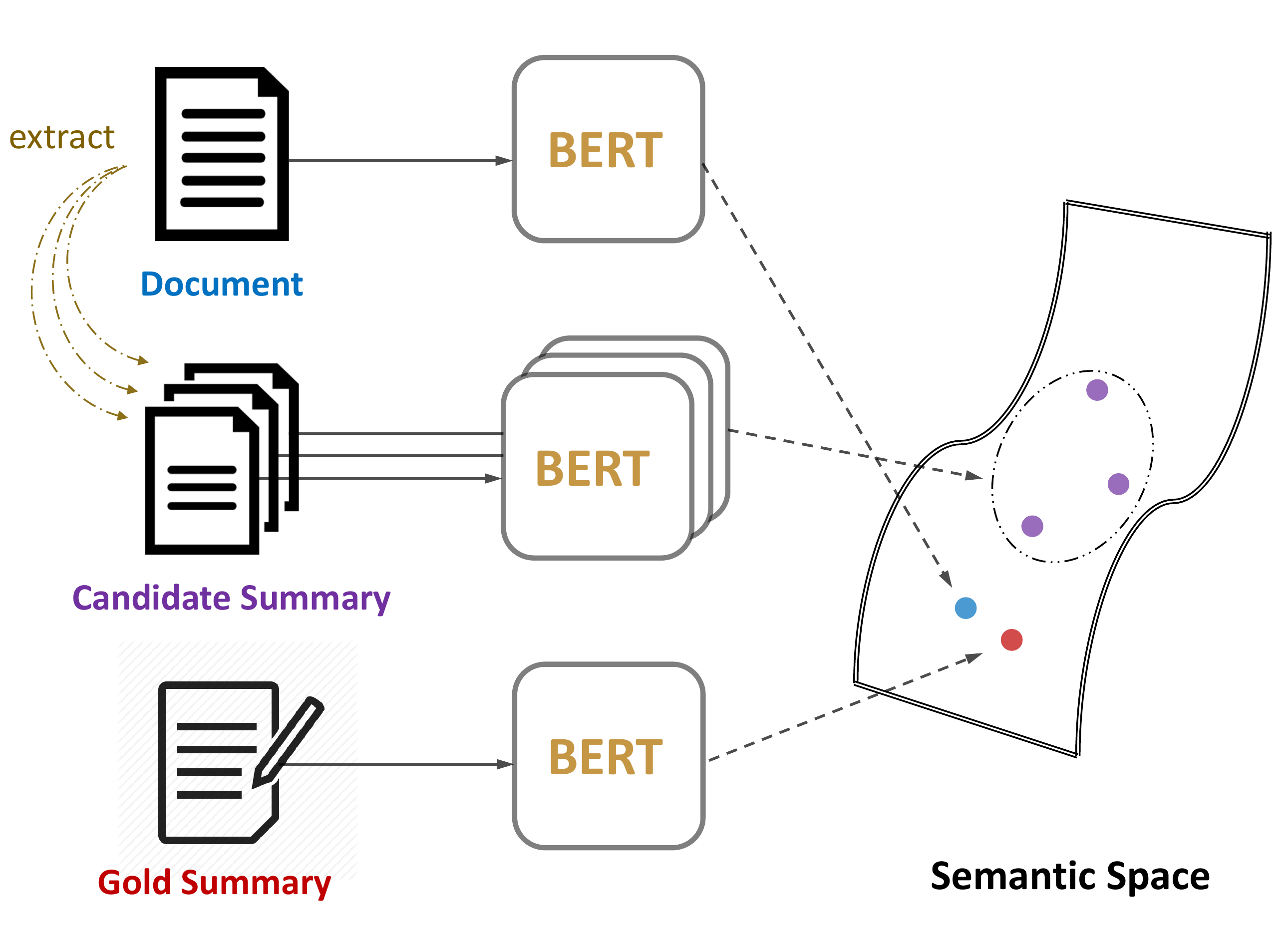}
    \caption{\textsc{MatchSum} framework. We match the contextual representations of the document with gold summary and candidate summaries (extracted from the document). Intuitively, better candidate summaries should be semantically closer to the document, while the gold summary should be the closest.}
    \label{fig:framework}
\end{figure}

The above systems of modeling the relationship between sentences are essentially sentence-level extractors, rather than considering the semantics of the entire summary. This makes them more inclined to select highly generalized sentences while ignoring the coupling of multiple sentences. \citet{narayan2018ranking,bae2019summary} utilize reinforcement learning (RL) to achieve summary-level scoring, but still limited to the architecture of sentence-level summarizers.

To better understand the advantages and limitations of sentence-level and summary-level approaches, we conduct an analysis on six benchmark datasets (in Section \ref{sec:investigation}) to explore the characteristics of these two methods. We find that there is indeed an inherent gap between the two approaches across these datasets, which motivates us to propose the following summary-level method.

In this paper, we propose a novel summary-level framework (\textsc{MatchSum}, Figure \ref{fig:framework}) and conceptualize extractive summarization as a semantic text matching problem. The principle idea is that a good summary should be more semantically similar as a whole to the source document than the unqualified summaries.
Semantic text matching is an important research problem to estimate semantic similarity between a source and a target text fragment, which has been applied in many fields, such as information retrieval \cite{mitra2017learning}, question answering \cite{yih2013question, severyn2015learning}, natural language inference \cite{wang2016learning, wang2017bilateral} and so on. One of the most conventional approaches to semantic text matching is to learn a vector representation for each text fragment, and then apply typical similarity metrics to compute the matching scores.

Specific to extractive summarization, we propose a Siamese-BERT architecture to compute the similarity between the source document and the candidate summary. Siamese BERT leverages the pre-trained
BERT \cite{devlin2019bert} in a Siamese network structure \cite{bromley1994signature, hoffer2015deep, reimers2019sentence} to derive semantically meaningful text embeddings that can be compared using cosine-similarity.
A good summary has the highest similarity among a set of candidate summaries.




We evaluate the proposed matching framework and perform significance testing on a range of benchmark datasets. Our model outperforms strong baselines significantly in all cases and improve the state-of-the-art extractive result on CNN/DailyMail. Besides, we design experiments to observe the gains brought by our framework.

We summarize our contributions as follows:

1) Instead of scoring and extracting sentences one by one to form a summary, we formulate extractive summarization as a semantic text matching problem and propose a novel summary-level framework. Our approach bypasses the difficulty of summary-level optimization by contrastive learning, that is, a good summary  should  be  more  semantically  similar to the source document than the unqualified summaries.


2) We conduct an analysis to investigate whether extractive models must do summary-level extraction based on the property of dataset, and attempt to quantify the inherent gap between sentence-level and summary-level methods.

3) Our proposed framework has achieved superior performance compared with strong baselines on six benchmark datasets. Notably, we obtain \textit{a state-of-the-art extractive result on CNN/DailyMail (44.41 in ROUGE-1) by only using the base version of BERT}. Moreover, we seek to observe where the performance gain of our model comes from.




\section{Related Work}

\subsection{Extractive Summarization}

Recent research work on extractive summarization spans a large range of approaches.
These work usually instantiate their encoder-decoder framework by choosing RNN \cite{zhou2018neural}, Transformer \cite{zhong2019closer, wang2019exploring} or GNN \cite{wang2020heterogeneous} as encoder, non-auto-regressive \cite{narayan2018ranking, arumae2018reinforced} or auto-regressive decoders \cite{jadhav2018extractive, liu2019text}. Despite the effectiveness, these models are essentially sentence-level extractors with individual scoring process favor the highest scoring sentence, which probably is not the optimal one to form summary\footnote{We will quantify this phenomenon in Section \ref{sec:investigation}.}.

The application of RL provides a means of summary-level scoring and brings improvement \cite{narayan2018ranking, bae2019summary}. However, these efforts are still limited to auto-regressive or non-auto-regressive architectures. Besides, in the non-neural approaches, the Integer Linear Programming (ILP) method can also be used for summary-level scoring \cite{wan2015multi}.

In addition, there is some work to solve extractive summarization from a semantic perspective before this paper, such as concept coverage \cite{gillick2009scalable}, reconstruction \cite{miao2016language} and maximize semantic volume \cite{yogatama2015extractive}.

\subsection{Two-stage Summarization}
Recent studies \cite{alyguliyev2009two, galanis2010extractive, zhang2019pretraining} have attempted to build two-stage document summarization systems. Specific to extractive summarization, the first stage is usually to extract some fragments of the original text, and the second stage is to select or modify on the basis of these fragments.



\citet{chen2018fast} and \citet{bae2019summary} follow a hybrid \textit{extract-then-rewrite} architecture, with policy-based RL to bridge the two networks together.
\citet{lebanoff2019scoring,xu-durrett-2019-neural,mendes2019jointly} focus on the \textit{extract-then-compress} learning paradigm, namely compressive summarization, which will first train an extractor for content selection.
Our model can be viewed as an \textit{extract-then-match} framework, which also employs a  sentence extractor to prune unnecessary information.






\section{Sentence-Level or Summary-Level? A Dataset-dependent Analysis}
\label{sec:investigation}

Although previous work has pointed out the weakness of sentence-level extractors, there is no systematic analysis towards the following questions:
1) For extractive summarization, is the \emph{summary-level extractor} better than the \emph{sentence-level extractor}?
2) Given a dataset, which extractor should we choose based on the characteristics of the data, and what is the inherent gap between these two extractors?

In this section, we investigate the gap between sentence-level and summary-level methods on six benchmark datasets, which can instruct us to search for an effective learning framework. It is worth noting that the sentence-level extractor we use here doesn't include a redundancy removal process so that we can estimate the effect of the summary-level extractor on redundancy elimination.
Notably, the analysis method to estimate the theoretical effectiveness presented in this section is generalized and can be applicable to any summary-level approach.

\renewcommand\arraystretch{1.1}
\begin{table*}[t]
    \center \footnotesize
    \tabcolsep0.13 in
    \begin{tabular}{lllcccccc}
    \toprule
    \multicolumn{1}{l}{\multirow{2}[1]{*}{\textbf{Datasets}}} &
    \multicolumn{1}{l}{\multirow{2}[1]{*}{\textbf{Source}}} &
    \multicolumn{1}{c}{\multirow{2}[1]{*}{\textbf{Type}}} & \multicolumn{3}{c}{\textbf{\# Pairs}} &
    \multicolumn{2}{c}{\textbf{\# Tokens}} &
    \multicolumn{1}{l}{\multirow{2}[1]{*}{\textbf{\# Ext}}} \\
     & & & Train & Valid & Test & Doc. & Sum. & \\
    \midrule
    Reddit &  Social Media & SDS & 41,675 & 645 & 645 &  482.2 & 28.0 & 2\\
    XSum  &  News  & SDS & 203,028 & 11,273 & 11,332 & 430.2 & 23.3 & 2\\
    CNN/DM &  News  & SDS & 287,084 & 13,367 & 11,489 & 766.1 & 58.2 & 3 \\
    WikiHow & Knowledge Base & SDS &  168,126 & 6,000 & 6,000 & 580.8 & 62.6 & 4\\
    PubMed &  Scientific Paper & SDS & 83,233 & 4,946 & 5,025 & 444.0 & 209.5 & 6\\
    Multi-News &  News  & MDS & 44,972 & 5,622 & 5,622 & 487.3 & 262.0 & 9\\
    \bottomrule
    \end{tabular}%
    \caption{Datasets overview. SDS represents single-document summarization and MDS represents multi-document summarization. The data in Doc. and Sum. indicates the average length of document and summary in the test set respectively. \# Ext denotes the number of sentences should extract in different datasets.}
  \label{tab:datasets}%
\end{table*}%

\subsection{Definition}
We refer to $D = \{s_1,\cdots,s_n\}$ as a single document consisting of $n$ sentences, and $C = \{s_1,\cdots,s_k, | s_i \in D\}$ as a \textit{candidate summary} including $k$ ($k \leq n$) sentences extracted from a document.
Given a document $D$ with its \textit{gold summary} $C^*$, we measure a \textit{candidate summary}  $C$  by calculating the ROUGE \cite{lin2003automatic} value between $C$  and $C^*$ in two levels:

1) Sentence-Level Score:
\begin{align}
    \mathrm{g}^{sen}(C) &= \frac{1}{|C|}\sum_{s \in C}\mathrm{R(s, C^*)} \label{eq:g_sen},
\end{align}
where $s$ is the sentence in $C$ and $|C|$ represents the number of sentences. $\mathrm{R}(\cdot)$ denotes the average ROUGE score\footnote{Here we use mean $\rm F_1$ of ROUGE-1, ROUGE-2 and ROUGE-L.}.
Thus, $\mathrm{g}^{sen}(C)$ indicates the average overlaps between each sentence in $C$ and the gold summary $C^*$.

2) Summary-Level Score:
\begin{align}
    \mathrm{g}^{sum}(C)  &=  \mathrm{R}(C, C^*) \label{eq:g_set},
\end{align}
where $\mathrm{g}^{sum}(C)$ considers sentences in $C$ as a whole and then calculates the ROUGE score with the gold summary $C^*$.


\paragraph{Pearl-Summary}
We define the \textit{pearl-summary} to be the summary that has a lower sentence-level score but a higher summary-level score.
\begin{Def}
A candidate summary $C$ is defined as a \textbf{pearl-summary} if there exists another candidate summary $C'$ that satisfies the inequality:
$
    \mathrm{g}^{sen}(C') > \mathrm{g}^{sen}(C)
$ while
$
    \mathrm{g}^{sum}(C') < \mathrm{g}^{sum}(C).
$
\end{Def}
Clearly, if a candidate summary is a pearl-summary, it is challenging for sentence-level summarizers to extract it.

\paragraph{Best-Summary}

The best-summary refers to a summary has highest summary-level score among all the candidate summaries.
\begin{Def}
A summary $\hat{C}$ is defined as the \textbf{best-summary} when it satisfies:
$\hat{C} = \mathop{\mathrm{argmax}}\limits_{C \in \mathcal{C}} \mathrm{g}^{sum} (C)$, where $\mathcal{C}$ denotes all the candidate summaries of the document.
\end{Def}

\subsection{Ranking of Best-Summary}
\label{sec:ranking}

For each document, we sort all candidate summaries\footnote{We use an approximate method here: take \#Ext (see Table \ref{tab:datasets}) of ten highest-scoring sentences to form candidate summaries.} in descending order based on the \textit{sentence-level score},
and then define $z$ as the rank index of the best-summary $\hat{C}$.

\begin{figure}[ht!]
  \centering
    \subfigure[Reddit]{
    \label{fig:reddit}
    \includegraphics[width=0.22\textwidth]{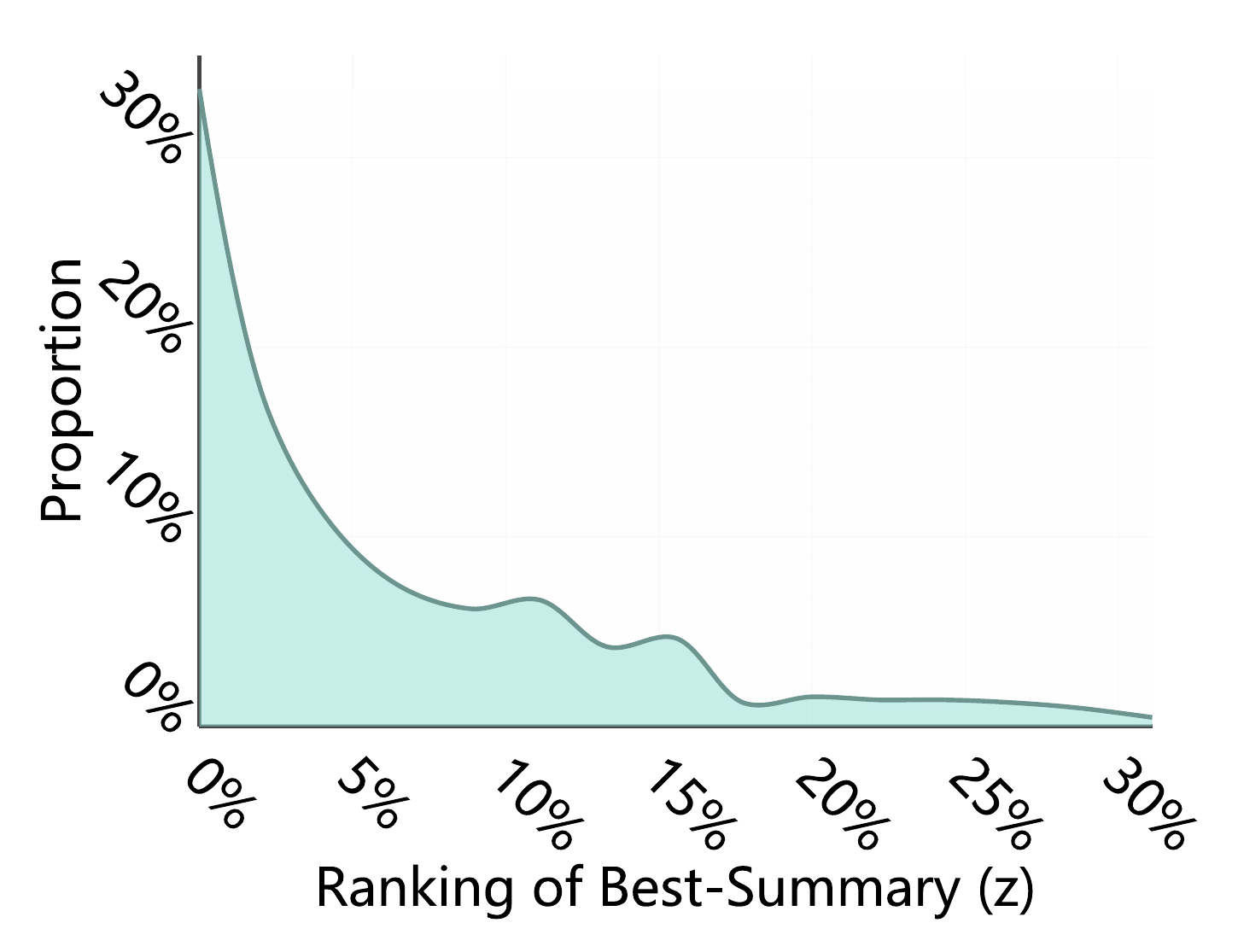}
  }
  \subfigure[XSum]{
    \label{fig:xsum}
    \includegraphics[width=0.22\textwidth]{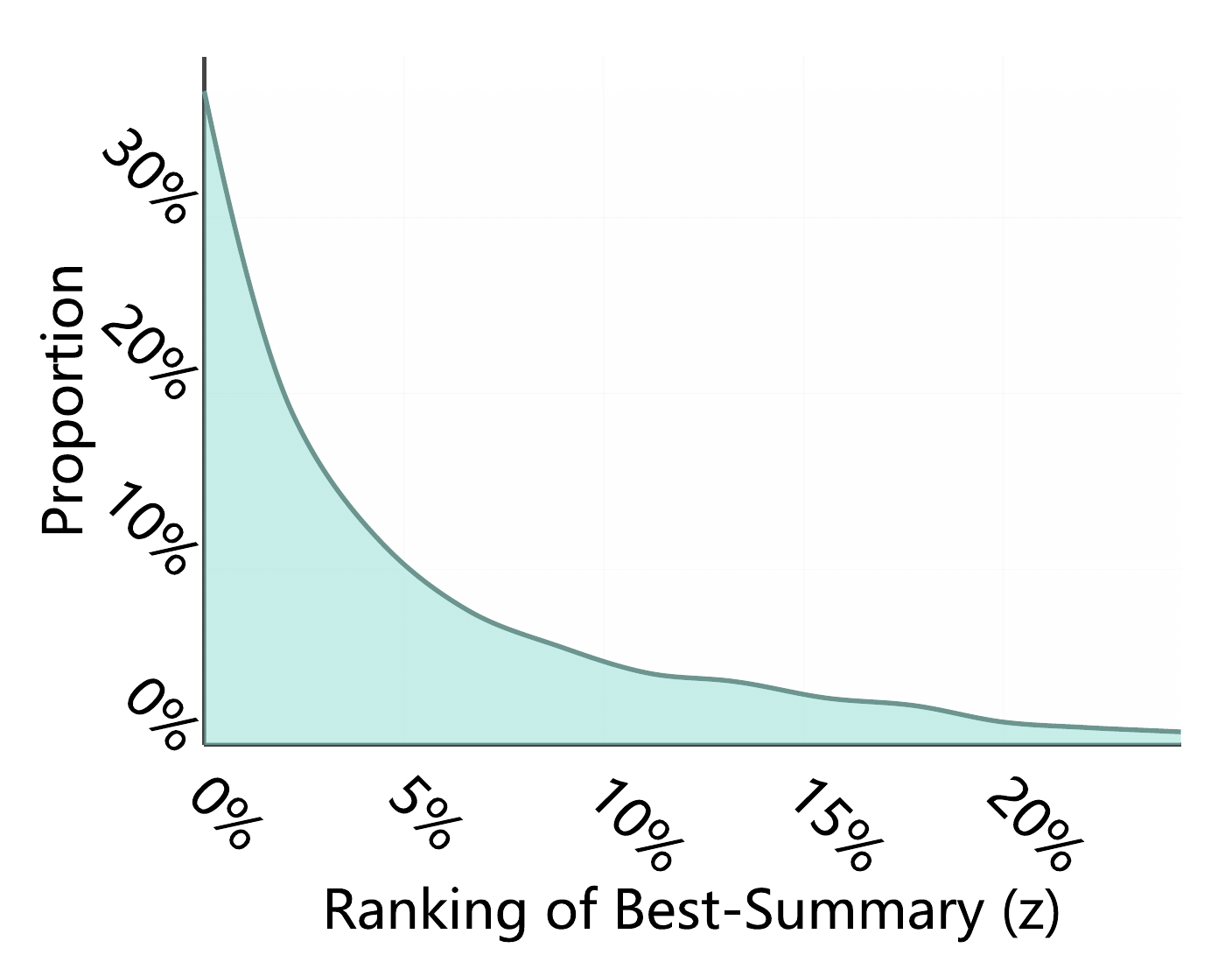}
  }
  \subfigure[CNN/DM]{
    \label{fig:cnndm}
    \includegraphics[width=0.22\textwidth]{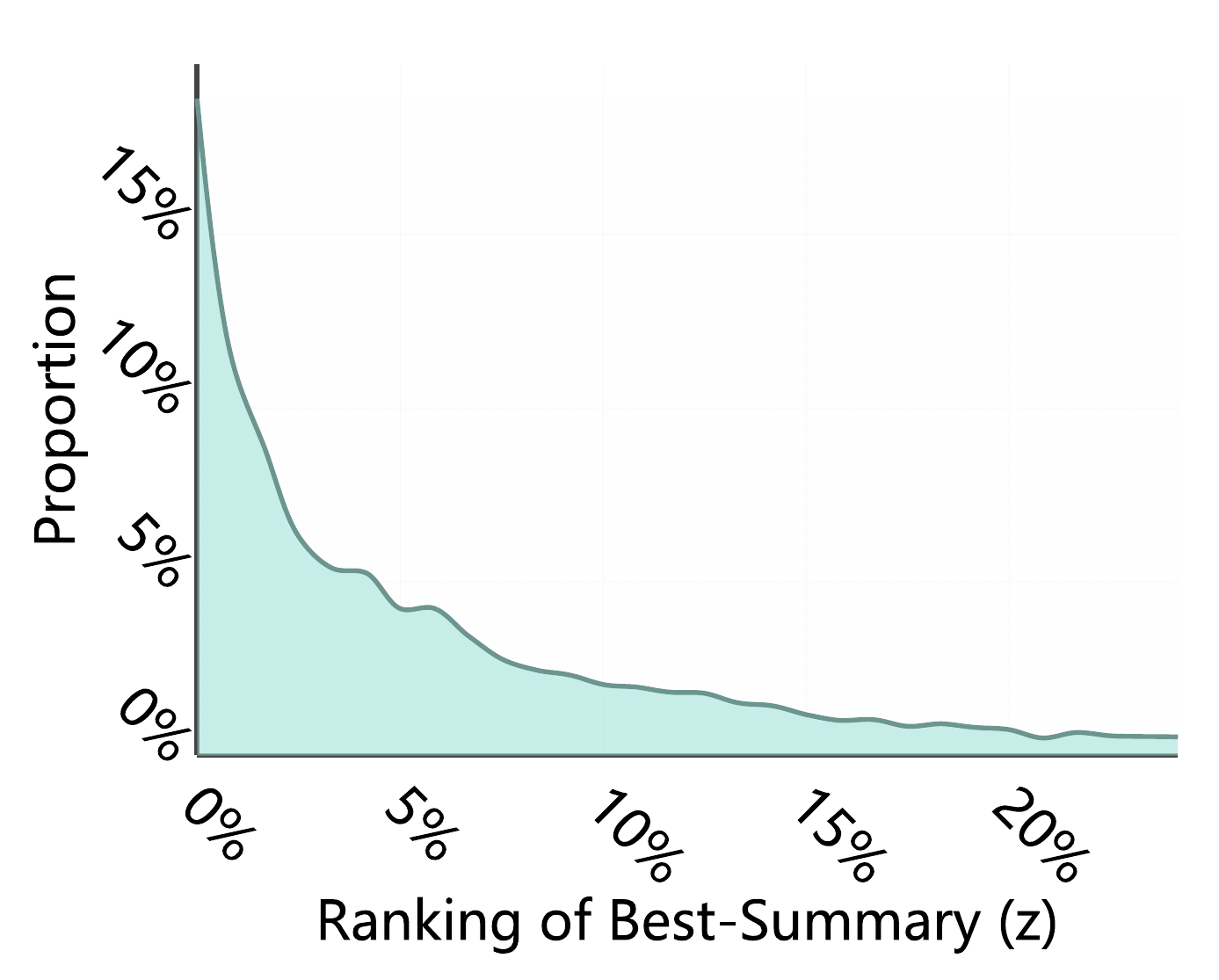}
  }
    \subfigure[WikiHow]{
    \label{fig:wikihow}
    \includegraphics[width=0.22\textwidth]{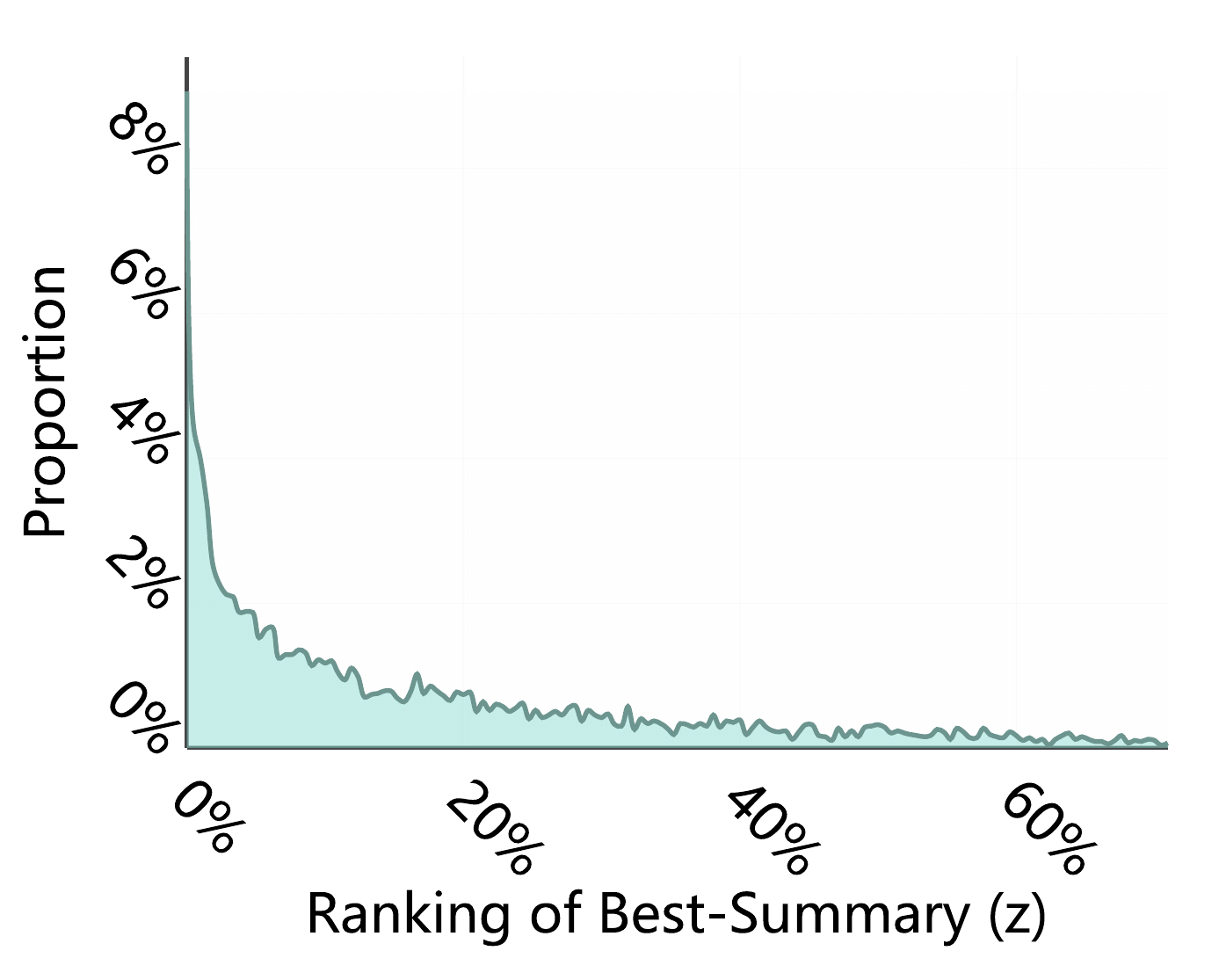}
  }
  \subfigure[PubMed]{
    \label{fig:pubmed}
    \includegraphics[width=0.22\textwidth]{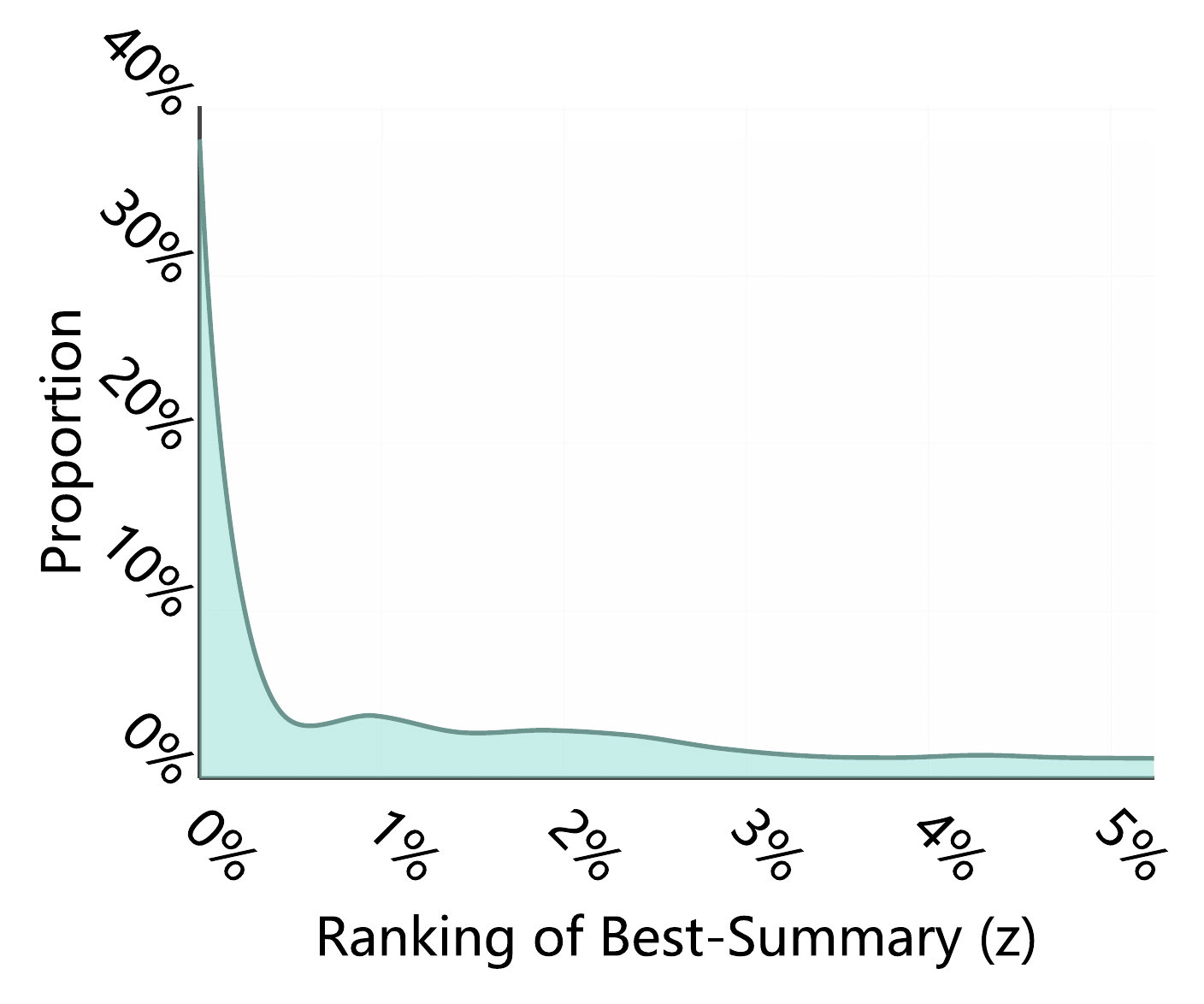}
  }
  \subfigure[Multi-News]{
    \label{fig:multinews}
    \includegraphics[width=0.22\textwidth]{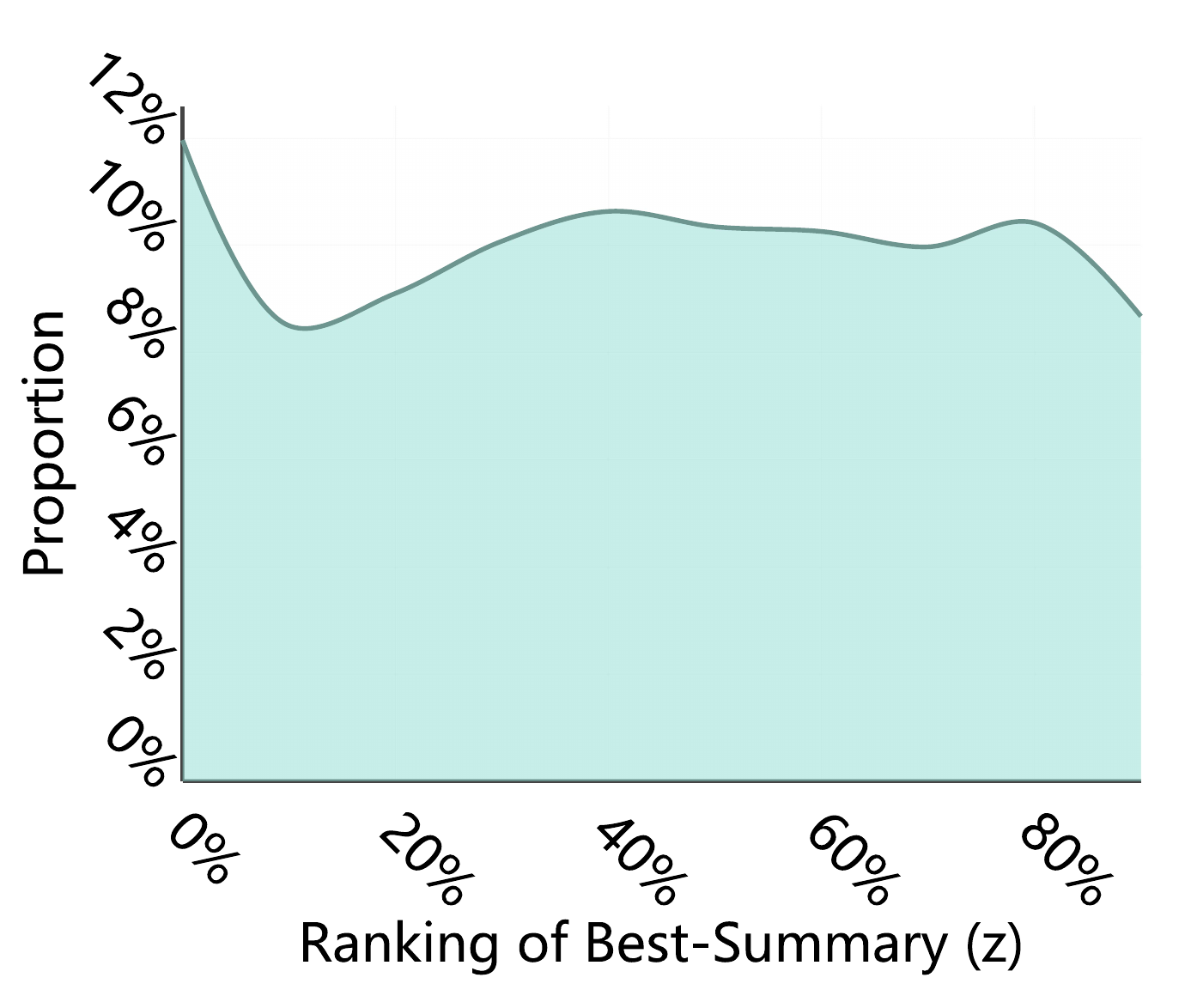}
  }
 \caption{Distribution of $z(\%)$ on six datasets. Because the number of candidate summaries for each document is different (short text may have relatively few candidates), we use $z$ / \textit{number of candidate summaries} as the X-axis. The Y-axis represents the proportion of the best-summaries with this rank in the test set.}
 \label{fig:dataset_fusion}
\end{figure}

Intuitively,
1) if $z=1$ ($\hat{C}$ comes first), it means that the best-summary is composed of sentences with the highest score;
2) If $z>1$, then the best-summary is a pearl-summary. And as $z$ increases ($\hat{C}$ gets lower rankings), we could find more candidate summaries whose sentence-level score is higher than best-summary, which leads to the learning difficulty for sentence-level extractors.

Since the appearance of the pearl-summary will bring challenges to sentence-level extractors, we attempt to investigate the proportion of pearl-summary in different datasets on six benchmark datasets. A detailed description of these datasets is displayed in Table~\ref{tab:datasets}.


As demonstrated in Figure \ref{fig:dataset_fusion}, we can observe that for all datasets,  most of the best-summaries are not made up of the highest-scoring sentences.
Specifically, for \texttt{CNN/DM}, only 18.9\% of best-summaries are not pearl-summary, indicating sentence-level extractors will easily fall into a local optimization, missing better candidate summaries.

Different from \texttt{CNN/DM}, \texttt{PubMed} is most suitable for sentence-level summarizers, because most of best-summary sets are not pearl-summary.
Additionally, it is challenging to achieve good performance on \texttt{WikiHow} and \texttt{Multi-News} without a summary-level learning process, as these two datasets are most evenly distributed, that is, the appearance of pearl-summary makes the selection of the best-summary more complicated.

In conclusion, the proportion of the pearl-summaries in all the best-summaries is a property to characterize a dataset, which will affect our choices of summarization extractors.

\subsection{Inherent Gap between Sentence-Level and Summary-Level Extractors}
\label{sec:inherent gap}
Above analysis has explicated that the summary-level method is better than the sentence-level method because it can pick out pearl-summaries, but how much improvement can it bring given a specific dataset?

Based on the definition of Eq. \eqref{eq:g_sen} and \eqref{eq:g_set}, we can characterize the upper bound of the sentence-level and summary-level summarization systems for a document $D$ as:

\begin{align}
    \alpha^{sen}(D) &= \max_{C\in \mathcal{C}_D}\mathrm{g}^{sen}(C), \\
    \alpha^{sum}(D) &= \max_{C\in \mathcal{C}_D}\mathrm{g}^{sum}(C),
\end{align}
where $\mathcal{C}_D$ is the set of candidate summaries extracted from $D$.

Then, we quantify the potential gain for a document $D$  by calculating the difference between $\alpha^{sen}(D)$ and $\alpha^{sum}(D)$:
\begin{align}
    \Delta(D) = \alpha^{sum}(D) - \alpha^{sen}(D).
\end{align}
Finally, a dataset-level potential gain can be obtained as:
\begin{align}
    \Delta(\mathcal{D}) = \frac{1}{|\mathcal{D}|}\sum_{D\in \mathcal{D}}\Delta(D),
\end{align}
where $\mathcal{D}$ represents a specific dataset and $|\mathcal{D}|$ is the number of documents in this dataset.

\begin{figure}[t]
    \centering
    \includegraphics[width=0.8\linewidth]{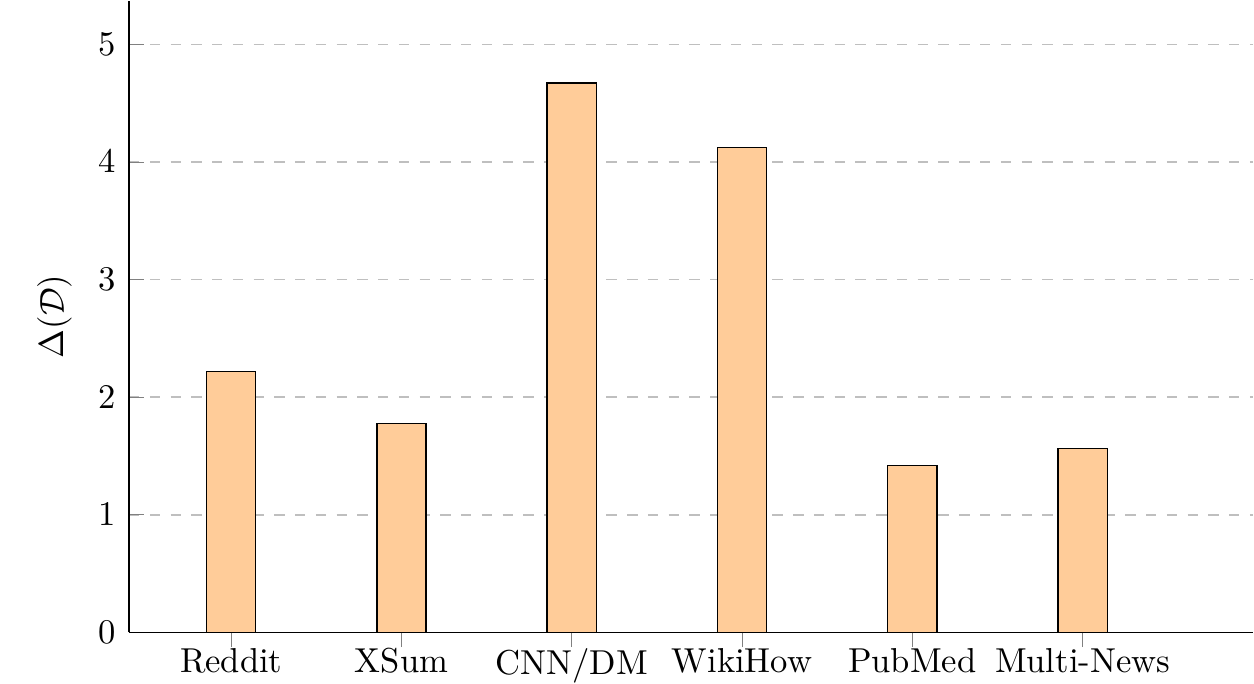}
    \caption{$\Delta(\mathcal{D})$ for different datasets.}
    \label{fig:delta}
\end{figure}

We can see from Figure \ref{fig:delta}, the performance gain of the summary-level method varies with the dataset and has an improvement at a maximum 4.7 on \texttt{CNN/DM}. From Figure \ref{fig:delta} and Table \ref{tab:datasets}, we can find the performance gain is related to the length of reference summary for different datasets. In the case of short summaries (\texttt{Reddit} and \texttt{XSum}), the perfect identification of pearl-summaries does not lead to much improvement. Similarly, multiple sentences in a long summary (\texttt{PubMed} and \texttt{Multi-News}) already have a large degree of semantic overlap, making the improvement of the summary-level method relatively small. But for a medium-length summary (\texttt{CNN/DM} and \texttt{WikiHow}, about 60 words), the summary-level learning process is rewarding. We will discuss this performance gain with specific models in Section \ref{sec:Analyzing}.

\section{Summarization as Matching}
\label{sec:model}






The above quantitative analysis suggests that for most of the datasets, sentence-level extractors are inherently unaware of pearl-summary, so obtaining the best-summary is difficult. To better utilize the above characteristics of the data, we propose a summary-level framework which could score and extract a summary directly.

Specifically, we formulate the extractive summarization task as a semantic text matching problem,   in  which  a  source  document and  candidate  summaries  will  be  (extracted from the original text) matched in a semantic space.
The following section will detail how we instantiate our proposed matching summarization framework by using a simple siamese-based architecture.

\subsection{Siamese-BERT}


Inspired by siamese network structure \cite{bromley1994signature}, we construct a Siamese-BERT architecture to match the document $D$ and the candidate summary $C$. Our Siamese-BERT consists of two BERTs with tied-weights and a cosine-similarity layer during the inference phase.

Unlike the modified BERT used in \cite{liu2019fine,bae2019summary}, we directly use the original BERT to derive the semantically meaningful embeddings from document $D$ and candidate summary $C$ since we need not obtain the sentence-level representation.
Thus, we use the vector of the `\texttt{[CLS]}' token from the top BERT layer as the representation of a document or summary.
Let $\mathbf{r}_D$ and $\mathbf{r}_C$ denote the embeddings of the document $D$ and candidate summary $C$. Their similarity score is measured by $f(D,C)=\mathrm{cosine}(\mathbf{r}_D,\mathbf{r}_C)$.

In order to fine-tune Siamese-BERT, we use a margin-based triplet loss to update the weights.
Intuitively, the gold summary $C^*$ should be semantically closest to the source document, which is the first principle our loss should follow:
\begin{align}
    \mathcal{L}_{1} = \max(0,f(D,C)-f(D,C^*) +\gamma_1),
\end{align}
where $C$ is the candidate summary in $D$ and $\gamma_1$ is a margin value. Besides, we also design a pairwise margin loss for all the candidate summaries. We sort all candidate summaries in descending order of ROUGE scores with the gold summary. Naturally, the candidate pair with a larger ranking gap should have a larger margin, which is the second principle to design our loss function:
\begin{equation}
\begin{aligned}
    \mathcal{L}_{2} = \max&(0, f(D,C_j)-f(D,C_i) \\
    &+(j-i)*\gamma_2) \quad  (i < j),
\end{aligned}
\end{equation}
where $C_i$ represents the candidate summary ranked $i$ and $\gamma_2$ is a hyperparameter used to distinguish between good and bad candidate summaries. Finally, our margin-based triplet loss can be written as:
\begin{align}
    \mathcal{L} = \mathcal{L}_{1} + \mathcal{L}_{2}.
\end{align}
The basic idea is to let the gold summary have the highest matching score, and at the same time,  a better candidate summary should obtain a higher score compared with the unqualified candidate summary. Figure \ref{fig:framework} illustrate this idea.


In the inference phase, we formulate extractive summarization as a task to search for the best summary among all the candidates $\mathcal{C}$ extracted from the document $D$.
\begin{align}
    \hat{C} = \mathop{\arg\,\max}_{C\in \mathcal{C}} f(D, C). \label{eq:search}
\end{align}









\subsection{Candidates Pruning}

\paragraph{Curse of Combination}
The matching idea is more intuitive while it suffers from combinatorial explosion problems. For example, how could we determine the size of the candidate summary set or should we score all possible candidates?
To alleviate these difficulties, we propose a simple candidate pruning strategy.

Concretely, we introduce a \textit{content selection module} to pre-select salient sentences.
The module learns to assign each sentence a salience score and prunes sentences irrelevant with the current document, resulting in a pruned document  ${D}^{'} = \{ s^{'}_1, \cdots, s^{'}_{ext} | s^{'}_i \in D \}$.

Similar to much previous work on two-stage summarization, our content selection module is a parameterized neural network. In this paper, we use \textsc{BertSum} \cite{liu2019text} without trigram blocking (we call it \textsc{BertExt}) to score each sentence. Then,  we use a simple rule to obtain the candidates: generating all combinations of $sel$ sentences subject to the pruned document, and reorganize the order of sentences according to the original position in the document to form candidate summaries. Therefore, we have a total of $\binom{ext}{sel}$ candidate sets.

\section{Experiment}

\subsection{Datasets}
In order to verify the effectiveness of our framework and obtain more convicing explanations, we perform experiments on six divergent mainstream datasets as follows.

\textbf{CNN/DailyMail} \cite{hermann2015teaching} is a commonly used summarization dataset modified by \citet{nallapati2016abstractive}, which contains news articles and
associated highlights as summaries. In this paper, we use the non-anonymized version.

\textbf{PubMed} \cite{cohan2018discourse} is collected from scientific papers and thus consists of long documents. We modify this dataset by using the introduction section as the document and the abstract section as the corresponding summary.

\textbf{WikiHow} \cite{koupaee2018wikihow} is a diverse dataset extracted from an online knowledge base. Articles in it span a wide range of topics.

\textbf{XSum} \cite{narayan2018don} is a one-sentence summary dataset to answer the question ``What is the article about?''. All summaries are professionally written, typically by the authors of documents in this dataset.

\textbf{Multi-News} \cite{DBLP:conf/acl/FabbriLSLR19} is a multi-document news summarization dataset with a relatively long summary, we use the truncated version and concatenate the source documents as a single input in all experiments.

\textbf{Reddit} \cite{kim2019abstractive} is a highly abstractive dataset collected from social media platform. We only use the TIFU-long version of Reddit, which regards the body text of a post as the document and the TL;DR as the summary.

\renewcommand\arraystretch{1.3}
\begin{table}[t]\footnotesize\setlength{\tabcolsep}{2.3pt}
  \centering
    \begin{tabular}{lcccccc}
    \toprule
      & \textbf{Reddit} & \textbf{XSum} & \textbf{CNN/DM} & \textbf{Wiki} & \textbf{PubMed} & \textbf{M-News} \\
    \midrule
    \textbf{Ext} & 5 & 5 & 5 & 5 & 7 & 10 \\
    \textbf{Sel} & 1, 2 & 1, 2 & 2, 3 & 3, 4, 5 & 6 & 9 \\
    \textbf{Size} & 15 & 15 & 20 & 16 & 7 & 9 \\
    \bottomrule
    \end{tabular}%
  \caption{Details about the candidate summary for different datasets. \textit{Ext} denotes the number of sentences after we prune the original document, \textit{Sel} denotes the number of sentences to form a candidate summary and \textit{Size} is the number of final candidate summaries.}
  \label{tab:candidate size}
\end{table}%

\subsection{Implementation Details}
We use the base version of BERT to implement our models in all experiments. Adam optimizer \cite{kingma2014adam} with warming-up is used and our learning rate schedule follows \citet{vaswani2017attention} as:
\begin{align}
    \mathrm{lr} = \mathrm{2e^{-3} \cdot min(step^{-0.5}, step \cdot wm^{-1.5})},
\end{align}
where each step is a batch size of 32 and $wm$ denotes warmup steps of 10,000. We choose $\gamma_1=0$ and $\gamma_2=0.01$. When $\gamma_1 \textless 0.05$ and $0.005 \textless \gamma_2 \textless 0.05$ they have little effect on performance, otherwise they will cause performance degradation. We use the validation set to save three best checkpoints during training, and record the performance of the best checkpoints on the test set. Importantly, all the experimental results listed in this paper are the average of three runs. To obtain a Siamese-BERT model on \texttt{CNN/DM}, we use 8 Tesla-V100-16G GPUs for about 30 hours of training.

For datasets, we remove samples with empty document or summary and truncate the document to 512 tokens, therefore ORACLE in this paper is calculated on the truncated datasets. Details of candidate summary for the different datasets can be found in Table \ref{tab:candidate size}.

\subsection{Experimental Results}

\renewcommand\arraystretch{1.2}
\begin{table}[t]
\center \footnotesize
\tabcolsep0.07in
\setlength{\tabcolsep}{1mm}{
\setlength{\tabcolsep}{1.4mm}{
\begin{tabular}{lccc}
\toprule
{\textbf{Model}} & \textbf{R-1} & \textbf{R-2} & \textbf{R-L}  \\
\midrule
LEAD & 40.43 & 17.62 & 36.67 \\
ORACLE & 52.59 & 31.23 & 48.87 \\
MATCH-ORACLE & 51.08 & 26.94 & 47.22 \\
\midrule
\textsc{BanditSum} \cite{dong2018banditsum} & 41.50 & 18.70 & 37.60 \\
\textsc{NeuSum} \cite{zhou2018neural} & 41.59 & 19.01 & 37.98 \\
\textsc{Jecs} \cite{xu-durrett-2019-neural} & 41.70 & 18.50 & 37.90 \\
\textsc{HiBert} \cite{Zhang2019HIBERTDL} & 42.37 & 19.95 & 38.83 \\
\textsc{PnBert} \cite{zhong2019searching} & 42.39 & 19.51 & 38.69 \\
\textsc{PnBert} + RL & 42.69 & 19.60 & 38.85 \\
\textsc{BertExt$^\dag$} \cite{bae2019summary} & 42.29 & 19.38 & 38.63 \\
\textsc{BertExt$^\dag$} + RL  & 42.76 & 19.87 & 39.11 \\
\textsc{BertExt} \cite{liu2019fine} & 42.57 & 19.96 & 39.04 \\
\textsc{BertExt} + Tri-Blocking & 43.23 & 20.22 & 39.60 \\
$\textsc{BertSum}^*$ \cite{liu2019text} & 43.85 & 20.34 & 39.90 \\

\midrule
\textsc{BertExt} (Ours) & 42.73 & 20.13 & 39.20 \\
\textsc{BertExt} + Tri-Blocking (Ours) & 43.18 & 20.16 & 39.56 \\
\textsc{MatchSum} (BERT-base) & 44.22 & 20.62 & 40.38 \\
\textsc{MatchSum} (RoBERTa-base) & \textbf{44.41} & \textbf{20.86} & \textbf{40.55} \\
\bottomrule
\end{tabular}}}
\caption{Results on CNN/DM test set. The model with $^*$ indicates that the large version of BERT is used. \textsc{BertExt$^\dag$} add an additional Pointer Network compared to other \textsc{BertExt} in this table.}
\label{table:cnndm}
\end{table}

\paragraph{Results on CNN/DM}
As shown in Table \ref{table:cnndm}, we list strong baselines with different learning approaches.
The first section contains \textit{LEAD}, \textit{ORACLE} and \textit{MATCH-ORACLE}\footnote{\textit{LEAD} and \textit{ORACLE} are common baselines in the summarization task. The former means extracting the first several sentences of a document as a summary, the latter is the groundtruth used in extractive models training. \textit{MATCH-ORACLE} is the groundtruth used to train \textsc{MatchSum}.}. Because we prune documents before matching, \textit{MATCH-ORACLE} is relatively low.

We can see from the second section, although RL can score the entire summary, it does not lead to much performance improvement. This is probably because it still relies on the sentence-level summarizers such as Pointer network or sequence labeling models, which select sentences one by one, rather than distinguishing the semantics of different summaries as a whole. Trigram Blocking is a simple yet effective heuristic on CNN/DM, even better than all redundancy removal methods based on neural models.

Compared with these models, our proposed \textsc{MatchSum} has outperformed all competitors by a large margin. For example, it beats \textsc{BertExt} by 1.51 ROUGE-1 score when using BERT-base as the encoder. Additionally, even compared with the baseline with BERT-large pre-trained encoder, our model \textsc{MatchSum} (BERT-base) still perform better. Furthermore, when we change the encoder to RoBERTa-base \cite{liu2019roberta}, the performance can be further improved. We think the improvement here is because RoBERTa introduced 63 million English news articles during pretraining. The superior performance on this dataset demonstrates the effectiveness of our proposed matching framework.

\renewcommand\arraystretch{1.2}
\begin{table}[t]
\center \footnotesize
\tabcolsep0.13 in
\begin{tabular}{lccc}
\toprule
\multicolumn{1}{c}{\textbf{Model}} & \textbf{R-1} & \textbf{R-2} & \textbf{R-L} \\
\midrule
\multicolumn{4}{c}{\textbf{Reddit}} \\
\midrule
\textsc{BertExt} (Num = 1) & 21.99 & 5.21	& 16.99 \\
\textsc{BertExt} (Num = 2) & 23.86 & 5.85	& 19.11 \\
\textsc{MatchSum} (Sel = 1) & 22.87 & 5.15 & 17.40 \\
\textsc{MatchSum} (Sel = 2) & 24.90 & 5.91 & 20.03 \\
\textsc{MatchSum} (Sel = 1, 2) & \textbf{25.09} & \textbf{6.17} & \textbf{20.13} \\

\midrule
\multicolumn{4}{c}{\textbf{XSum}} \\
\midrule
\textsc{BertExt} (Num = 1) & 22.53 & 4.36 & 16.23 \\
\textsc{BertExt} (Num = 2) & 22.86 & 4.48 & 17.16 \\
\textsc{MatchSum} (Sel = 1) & 23.35 & 4.46 & 16.71 \\
\textsc{MatchSum} (Sel = 2) & 24.48	& 4.58 & 18.31 \\
\textsc{MatchSum} (Sel = 1, 2) & \textbf{24.86} & \textbf{4.66} & \textbf{18.41} \\

\bottomrule
\end{tabular}
\caption{Results on test sets of Reddit and XSum. $Num$ indicates how many sentences \textsc{BertExt} extracts as a summary and $Sel$ indicates the number of sentences we choose to form a candidate summary.} \label{tab:abstractive datasets}
\end{table}

\renewcommand\arraystretch{1.1}
\begin{table*}[t]
\center \footnotesize
\tabcolsep0.13 in
\begin{tabular}{lccccccccc}
\toprule
\multicolumn{1}{c}{\multirow{2}[1]{*}{\textbf{Model}}}  &
\multicolumn{3}{c}{\textbf{WikiHow}} &
\multicolumn{3}{c}{\textbf{PubMed}} &
\multicolumn{3}{c}{\textbf{Multi-News}} \\

 & \textbf{R-1} & \textbf{R-2} & \textbf{R-L} &
\textbf{R-1} & \textbf{R-2} & \textbf{R-L} &
\textbf{R-1} & \textbf{R-2} & \textbf{R-L} \\

\cmidrule(lr){1-1} \cmidrule(lr){2-4} \cmidrule(lr){5-7} \cmidrule(lr){8-10}

LEAD & 24.97 & 5.83 & 23.24 & 37.58 & 12.22 & 33.44 & 43.08 & 14.27 & 38.97 \\
ORACLE & 35.59 & 12.98 & 32.68 & 45.12 & 20.33 & 40.19 & 49.06 & 21.54 & 44.27  \\
MATCH-ORACLE & 35.22 & 10.55 & 32.87 & 42.21 & 15.42 & 37.67 & 47.45 & 17.41 & 43.14 \\

\cmidrule(lr){1-1} \cmidrule(lr){2-4} \cmidrule(lr){5-7} \cmidrule(lr){8-10}
\textsc{BertExt} & 30.31 & 8.71	& 28.24 & 41.05 & 14.88	& 36.57 & 45.80 & 16.42 & 41.53 \\
\quad + 3gram-Blocking & 30.37 & 8.45 & 28.28 & 38.81 & 13.62 & 34.52 & 44.94 & 15.47 & 40.63 \\
\quad + 4gram-Blocking & 30.40 & 8.67 & 28.32 & 40.29 & 14.37 & 35.88 & 45.86 & 16.23 & 41.57 \\
\textsc{MatchSum} (BERT-base) & \textbf{31.85} & \textbf{8.98} & \textbf{29.58} & \textbf{41.21} & \textbf{14.91} & \textbf{36.75} & \textbf{46.20} & \textbf{16.51} & \textbf{41.89} \\

\bottomrule
\end{tabular}
\caption{Results on test sets of WikiHow, PubMed and Multi-News. \textsc{MatchSum} beats the state-of-the-art BERT model with Ngram Blocking on all different domain datasets.
}
\label{tab:ngram and match}
\end{table*}

\paragraph{Results on Datasets with Short Summaries}
\texttt{Reddit} and \texttt{XSum} have been heavily evaluated by abstractive summarizer due to their short summaries.
Here, we evaluate our model on these two datasets to investigate whether \textsc{MatchSum} could achieve improvement when dealing with summaries containing fewer sentences compared with other typical extractive models.


When taking just one sentence to match the original document, \textsc{MatchSum} degenerates into a re-ranking of sentences. Table \ref{tab:abstractive datasets} illustrates that this degradation can still bring a small improvement (compared to \textsc{BertExt} (Num = 1),  0.88 $\Delta$R-1 on \texttt{Reddit}, 0.82 $\Delta$R-1 on \texttt{XSum}). However, when the number of sentences increases to two and summary-level semantics need to be taken into account, \textsc{MatchSum} can obtain a more remarkable improvement (compared to \textsc{BertExt} (Num = 2), 1.04 $\Delta$R-1 on \texttt{Reddit}, 1.62 $\Delta$R-1 on \texttt{XSum}).

In addition, our model maps candidate summary as a whole into semantic space, so it can flexibly choose any number of sentences, while most other methods can only extract a fixed number of sentences. From Table \ref{tab:abstractive datasets}, we can see this advantage leads to further performance improvement.

\paragraph{Results on Datasets with Long Summaries}
When the summary is relatively long, summary-level matching becomes more complicated and is harder to learn.
We aim to compare the difference between Trigram Blocking and our model when dealing with long summaries.


Table \ref{tab:ngram and match} presents that although Trigram Blocking works well on \texttt{CNN/DM}, it does not always maintain a stable improvement. Ngram Blocking has little effect on \texttt{WikiHow} and \texttt{Multi-News}, and it causes a large performance drop on \texttt{PubMed}. We think the reason is that Ngram Blocking cannot really understand the semantics of sentences or summaries, just restricts the presence of entities with many words to only once, which is obviously not suitable for the scientific domain where entities may often appear multiple times.

On the contrary, our proposed method does not have these strong constraints, but aligns the original document with the summary from semantic space. Experiment results display that our model is robust on all domains, especially on \texttt{WikiHow}, \textsc{MatchSum} beats the state-of-the-art BERT model by 1.54 ROUGE-1 score.

\begin{figure*}[t]
  \centering
  \subfigure[XSum]{
    \label{fig:xsum_split}
    \includegraphics[width=0.28\textwidth]{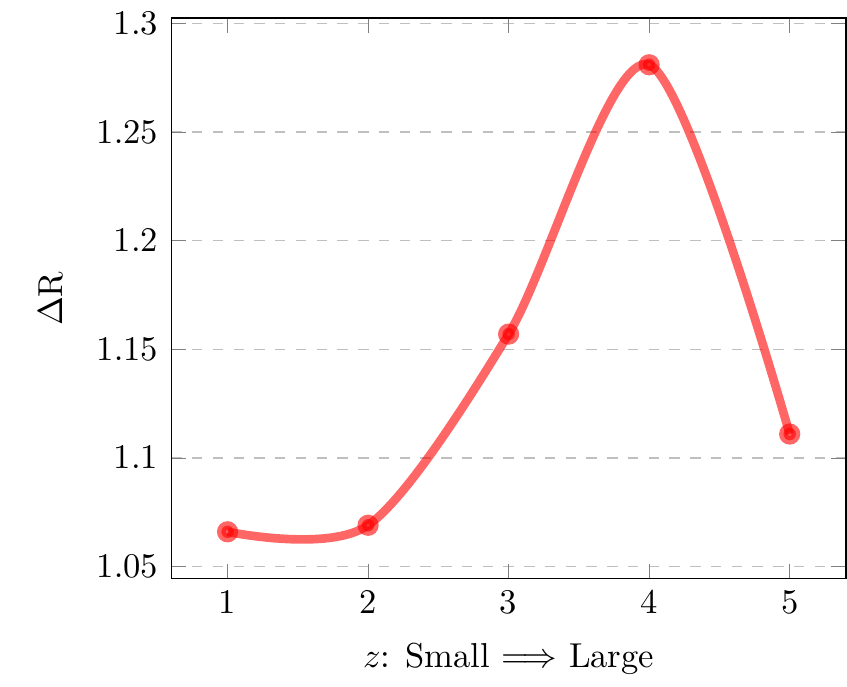}
  }
  \subfigure[CNN/DM]{
    \label{fig:cnndm_split}
    \includegraphics[width=0.28\textwidth]{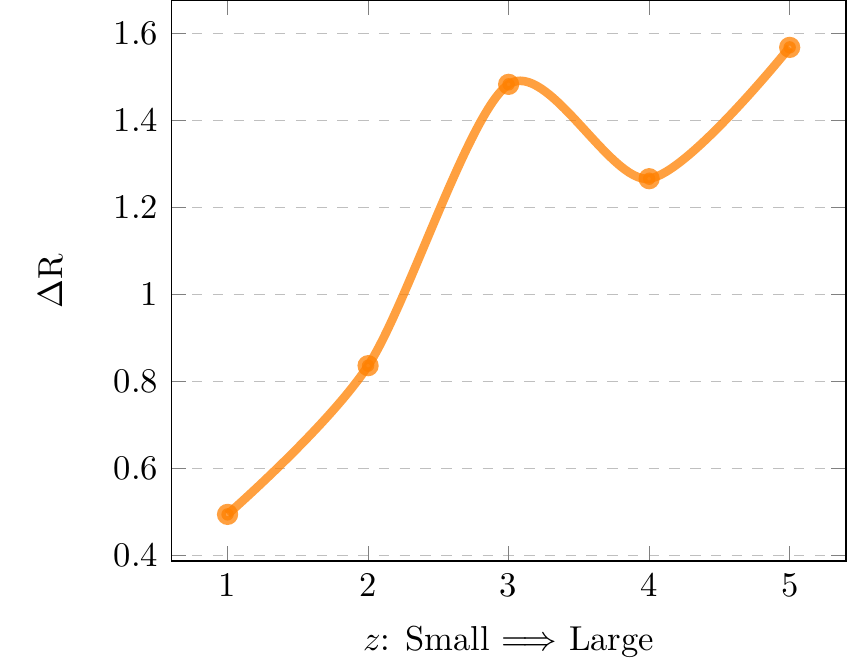}
  }
  \subfigure[WikiHow]{
    \label{fig:pubmed_split}
    \includegraphics[width=0.28\textwidth]{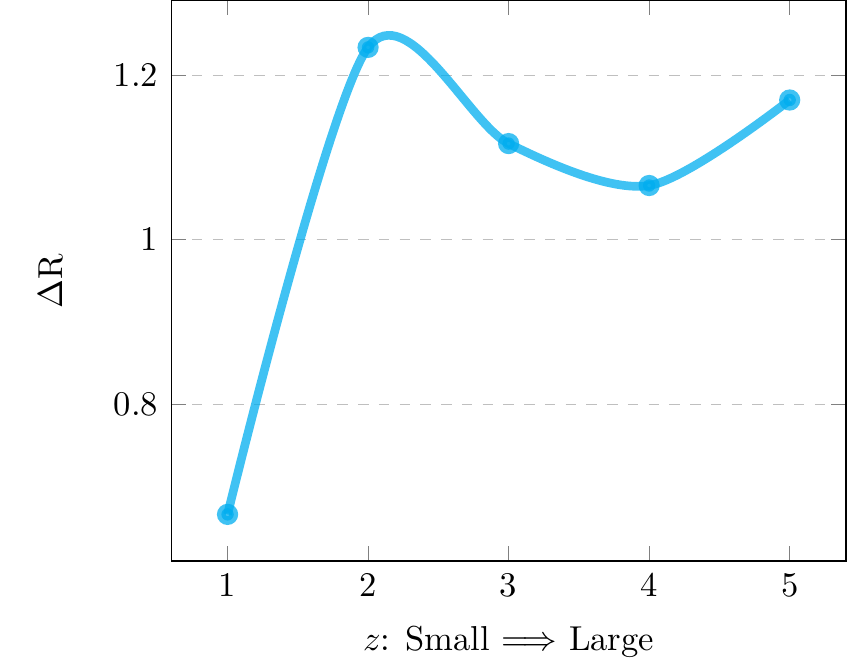}
  }
 \caption{Datasets splitting experiment. We split test sets into five parts according to $z$ described in Section \ref{sec:ranking}. The X-axis from left to right indicates the subsets of the test set with the value of $z$ from small to large, and the Y-axis represents the ROUGE improvement of \textsc{MatchSum} over \textsc{BertExt} on this subset.}
 \label{fig:split}
\end{figure*}

\begin{figure}
    \centering
    \includegraphics[width=0.7\linewidth]{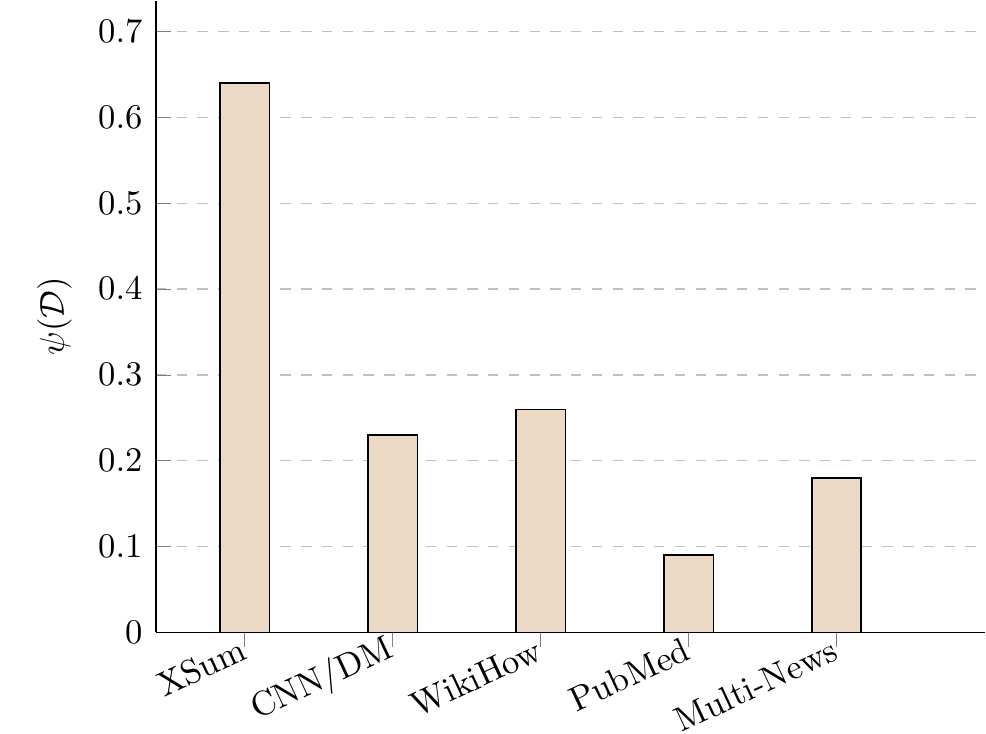}
    \caption{$\psi$ of different datasets. Reddit is excluded because it has too few samples in the test set.}
    \label{fig:ratio}
\end{figure}

\subsection{Analysis}
\label{sec:Analyzing}
In the following, our analysis is driven by two questions:

1) Whether the benefits of \textsc{MatchSum} are consistent with the property of the dataset analyzed in Section~\ref{sec:investigation}?

2) Why have our model achieved different performance gains on diverse datasets?


\paragraph{Dataset Splitting Testing} Typically, we choose three datasets (\texttt{XSum}, \texttt{CNN/DM} and \texttt{WikiHow}) with the largest performance gain for this experiment. We split each test set into roughly equal numbers of five parts according to $z$ described in Section \ref{sec:ranking}, and then experiment with each subset.

Figure \ref{fig:split} shows that the performance gap between \textsc{MatchSum} and \textsc{BertExt} is always the smallest when the best-summary is not a pearl-summary ($z = 1$). The phenomenon is in line with our understanding, in these samples, the ability of the summary-level extractor to discover pearl-summaries does not bring advantages.

As $z$ increases, the performance gap generally tends to increase. Specifically, the benefit of \textsc{MatchSum} on \texttt{CNN/DM} is highly consistent with the appearance of pearl-summary. It can only bring an improvement of 0.49 in the subset with the smallest $z$, but it rises sharply to 1.57 when $z$ reaches its maximum value. \texttt{WikiHow} is similar to \texttt{CNN/DM}, when best-summary consists entirely of highest-scoring sentences, the performance gap is obviously smaller than in other samples. \texttt{XSum} is slightly different, although the trend remains the same, our model does not perform well in the samples with the largest $z$, which needs further improvement and exploration.

From the above comparison, we can see that the performance improvement of \textsc{MatchSum} is concentrated in the samples with more pearl-summaries, which illustrates our semantic-based summary-level model can capture sentences that are not particularly good when viewed individually, thereby forming a better summary.

\paragraph{Comparison Across Datasets}
Intuitively, improvements brought by \textsc{MatchSum} framework should be associated with inherent gaps presented in Section \ref{sec:inherent gap}. To better understand their relation, we introduce $\Delta(\mathcal{D})^*$ as follows:
\begin{align}
    \Delta(D)^* &= \mathrm{g}^{sum}(C_{MS}) - \mathrm{g}^{sum}(C_{BE}), \\
    \Delta(\mathcal{D})^* &= \frac{1}{|\mathcal{D}|}\sum_{D\in \mathcal{D}}\Delta(D)^*,
\end{align}
where $C_{MS}$ and $C_{BE}$ represent the candidate summary selected by \textsc{MatchSum} and \textsc{BertExt} in the document $D$, respectively. Therefore, $\Delta(\mathcal{D})^*$ can indicate the improvement by \textsc{MatchSum} over \textsc{BertExt} on dataset $\mathcal{D}$. Moreover, compared with the inherent gap between sentence-level and summary-level extractors, we define the ratio that \textsc{MatchSum} can learn on dataset $\mathcal{D}$ as:
\begin{align}
    \psi(\mathcal{D}) = \Delta(\mathcal{D})^* / \Delta(\mathcal{D}),
\end{align}
where $\Delta(\mathcal{D})$ is the inherent gap between sentence-level and summary-level extractos.

It is clear from Figure \ref{fig:ratio}, the value of $\psi(\mathcal{D})$ depends on $z$ (see Figure \ref{fig:dataset_fusion}) and the length of the gold summary (see Table \ref{tab:datasets}). As the gold summaries get longer, the upper bound of summary-level approaches becomes more difficult for our model to reach. \textsc{MatchSum} can achieve 0.64 $\psi(\mathcal{D})$  on \texttt{XSum} (23.3 words summary), however, $\psi(\mathcal{D})$ is less than 0.2 in \texttt{PubMed} and \texttt{Multi-News} whose summary length exceeds 200. From another perspective, when the summary length are similar, our model performs better on datasets with more pearl-summaries. For instance, $z$ is evenly distributed in \texttt{Multi-News} (see Figure \ref{fig:dataset_fusion}), so higher $\psi(\mathcal{D})$ (0.18) can be obtained than \texttt{PubMed} (0.09), which has the least pearl-summaries.

A better understanding of the dataset allows us to get a clear awareness of the strengths and limitations of our framework, and we also hope that the above analysis could provide useful clues for future research on extractive summarization.

\section{Conclusion}
We formulate the extractive summarization task as a semantic text matching problem and propose a novel summary-level framework to match the source document and candidate summaries in the semantic space.
We conduct an analysis to show how our model could better fit the characteristic of the data. Experimental results show \textsc{MatchSum} outperforms the current state-of-the-art extractive model on six benchmark datasets, which demonstrates the effectiveness of our method. We believe the power of this matching-based summarization framework has not been fully exploited. In the future, more forms of matching models can be explored to instantiated the proposed framework.

\section*{Acknowledgment}
We would like to thank the anonymous reviewers for their valuable comments. This work is supported by the National Key Research and Development Program of China (No. 2018YFC0831103), National Natural Science Foundation of China (No. U1936214 and 61672162), Shanghai Municipal Science and Technology Major Project (No. 2018SHZDZX01) and ZJLab.

\bibliography{acl2020}
\bibliographystyle{acl_natbib}




\end{document}



\appendix

\section{Appendices}
\label{sec:appendix}

\subsection{Introduction to datasets} 
In this paper, we perform experiments on six widely divergent mainstream datasets as follows.

\paragraph{CNN/DailyMail} \cite{hermann2015teaching} is a commonly used summarization dataset modified by \citet{nallapati2016abstractive}, which contains news articles and
associated highlights as summaries. In this paper, we use the non-anonymized version.

\paragraph{PubMed} \cite{cohan2018discourse} is collected from scientific papers and therefore consists of long documents. In our experiment, we modify this dataset by using the introduction section as the main document and the abstract section as the corresponding summary.

\paragraph{WikiHow} \cite{koupaee2018wikihow} is a diverse dataset extracted from an online knowledge base. Articles in it span a wide range of topics.

\paragraph{XSum} \cite{narayan2018don} is a one-sentence summary dataset to answer the question ``What is the article about?''. All summaries are professionally written, typically by the authors of documents in this dataset.

\paragraph{Multi-News} \cite{DBLP:conf/acl/FabbriLSLR19} is a multi-document news summarization dataset with a relatively long summary, we use the truncated version and concatenate the source documents as a single input in all experiments.

\paragraph{Reddit} \cite{kim2019abstractive} is a highly abstractive dataset collected from social media platform. We only use the TIFU-long version of Reddit, which regards the body text of a post as the document and the TL;DR as the summary.

\subsection{Implementation Details}
We use the base version of BERT to implement our models in all experiments. Adam optimizer \cite{kingma2014adam} with warming-up is used and our learning rate schedule follows \citet{vaswani2017attention} as:
\begin{align}
    \mathrm{lr} = \mathrm{2e^{-3} \cdot min(step^{-0.5}, step \cdot wm^{-1.5})},
\end{align}
where each step is a batch size of 32 and $wm$ denotes warmup steps of 10,000. Besides, we choose $\gamma_1=0$ and $\gamma_2=0.01$.

For datasets, we remove samples with empty document or summary and truncate the document to 512 tokens, therefore ORACLE in this paper is calculated on the truncated datasets. Details of candidate summary for the different datasets can be found in Table \ref{tab:candidate size}.

\renewcommand\arraystretch{1.3}
\begin{table}[t]\footnotesize\setlength{\tabcolsep}{2.3pt}
  \centering
    \begin{tabular}{lcccccc}
    \toprule
      & \textbf{Reddit} & \textbf{XSum} & \textbf{CNN/DM} & \textbf{Wiki} & \textbf{PubMed} & \textbf{M-News} \\
    \midrule
    \textbf{Ext} & 5 & 5 & 5 & 5 & 7 & 10 \\
    \textbf{Sel} & 1, 2 & 1, 2 & 2, 3 & 3, 4, 5 & 6 & 9 \\
    \textbf{Size} & 15 & 15 & 20 & 16 & 7 & 9 \\
    \bottomrule
    \end{tabular}%
  \caption{Details about the candidate summary for different datasets. \textit{Ext} represents the number of sentences after we prune the original document, \textit{Sel} represents the number of sentences we choose to form a candidate summary and \textit{Size} is the number of final candidate summaries.}
  \label{tab:candidate size}
\end{table}%


\bibliography{acl2020}
\bibliographystyle{acl_natbib}